\documentclass{article}

\usepackage{times}
\usepackage{epsfig}
\usepackage{graphicx}
\usepackage{amsmath}
\usepackage{amssymb}
\usepackage{booktabs} % for professional tables
\usepackage{threeparttable} % for table footnotes
\usepackage{url}
\usepackage{float}
\usepackage{caption}
\usepackage[subrefformat=parens,labelformat=parens]{subcaption}
\usepackage{mathtools}
\usepackage[symbol]{footmisc}
\usepackage{amsmath}% for multiline* env.

\usepackage[numbers]{natbib}

\usepackage[accepted]{icml2020}
\usepackage[pagebackref=true,breaklinks=true,letterpaper=true,colorlinks,bookmarks=false]{hyperref}

\icmltitlerunning{Layer-Wise Quantization Analysis}

\begin{document}

% \twocolumn[
%  	\icmltitle{Exploring Neural Networks Quantization via Layer-Wise Quantization Analysis}
% 	\begin{icmlauthorlist}
% 	% TODO return affiliations when submitting to ICML
% 		\icmlauthor{Shachar Gluska}{hailo}
% 		\icmlauthor{Mark Grobman}{hailo}
% 	\end{icmlauthorlist}

% 	\icmlaffiliation{hailo}{Hailo Technologies, Tel-Aviv, Gush Dan, Israel}
% 	\icmlcorrespondingauthor{Shachar Gluska}{shacharg@hailo.ai}
% 	\icmlcorrespondingauthor{Mark Grobman}{markg@hailo.ai}

% 	\vskip 0.3in

% ]

% TODO UNCOMMENT WHEN SUBMITTING TO ICML
% \printAffiliationsAndNotice{}

\title{Exploring Neural Networks Quantization via Layer-Wise Quantization Analysis}
\author{Shachar Gluska \qquad Mark Grobman\\
Hailo Technologies\\\{shacharg, markg\}@hailo.ai}
\date{\vspace{-5ex}}
\maketitle
% delete numbering for this page
\thispagestyle{empty}

%%%%%%%%% ABSTRACT
\begin{abstract}
	Quantization is an essential step in the efficient deployment of deep learning models and as such is an increasingly popular research topic. An important practical aspect that is not addressed in the current literature is how to analyze and fix fail cases where the use of quantization results in excessive degradation. In this paper, we present a simple analytic framework that breaks down overall degradation to its per layer contributions.  We analyze many common networks and observe that a layer's contribution is determined by both intrinsic (local) factors -- the distribution of the layer's weights and activations -- and extrinsic (global) factors having to do with the the interaction with the rest of the layers. Layer-wise analysis of existing quantization schemes reveals local fail-cases of existing techniques which are not reflected when inspecting their overall performance. As an example, we consider ResNext26 on which SoTA post-training quantization methods perform poorly. We show that almost all of the degradation stems from a single layer. The same analysis also allows for local fixes -- applying a common weight clipping heuristic only to this layer reduces degradation to a minimum while applying the same heuristic globally results in high degradation. More generally, layer-wise analysis allows for a more nuanced examination of how quantization affects the network, enabling the design of better performing schemes.
\end{abstract}

\begin{figure}[!th]
	\begin{center}
		\vspace{0.75cm}
		\begin{subfigure}{0.5\textwidth}
            \includegraphics[width=1.0\linewidth]{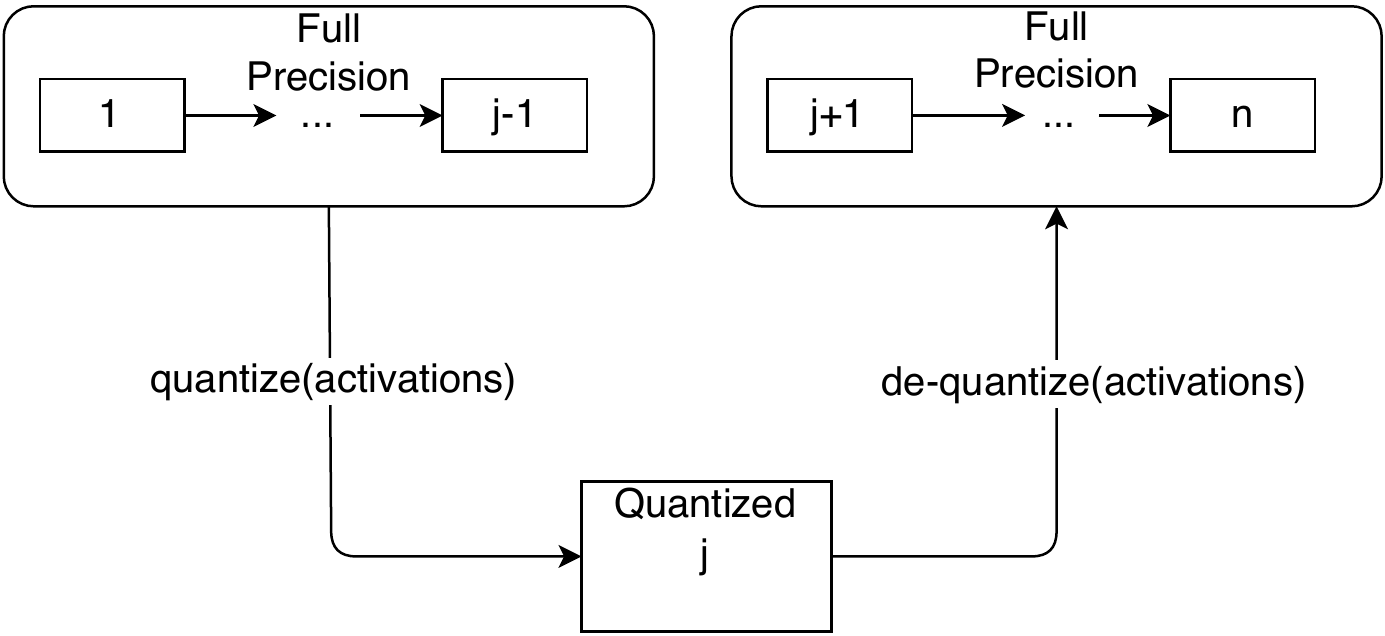}
            \caption{Basic building block of layer-wise analysis}
            \label{fig:anaylsis_building_block}
        \end{subfigure}
        \newline
        \vspace{0.25cm}
        \begin{subfigure}{0.5\textwidth}
            \includegraphics[width=1.0\linewidth]{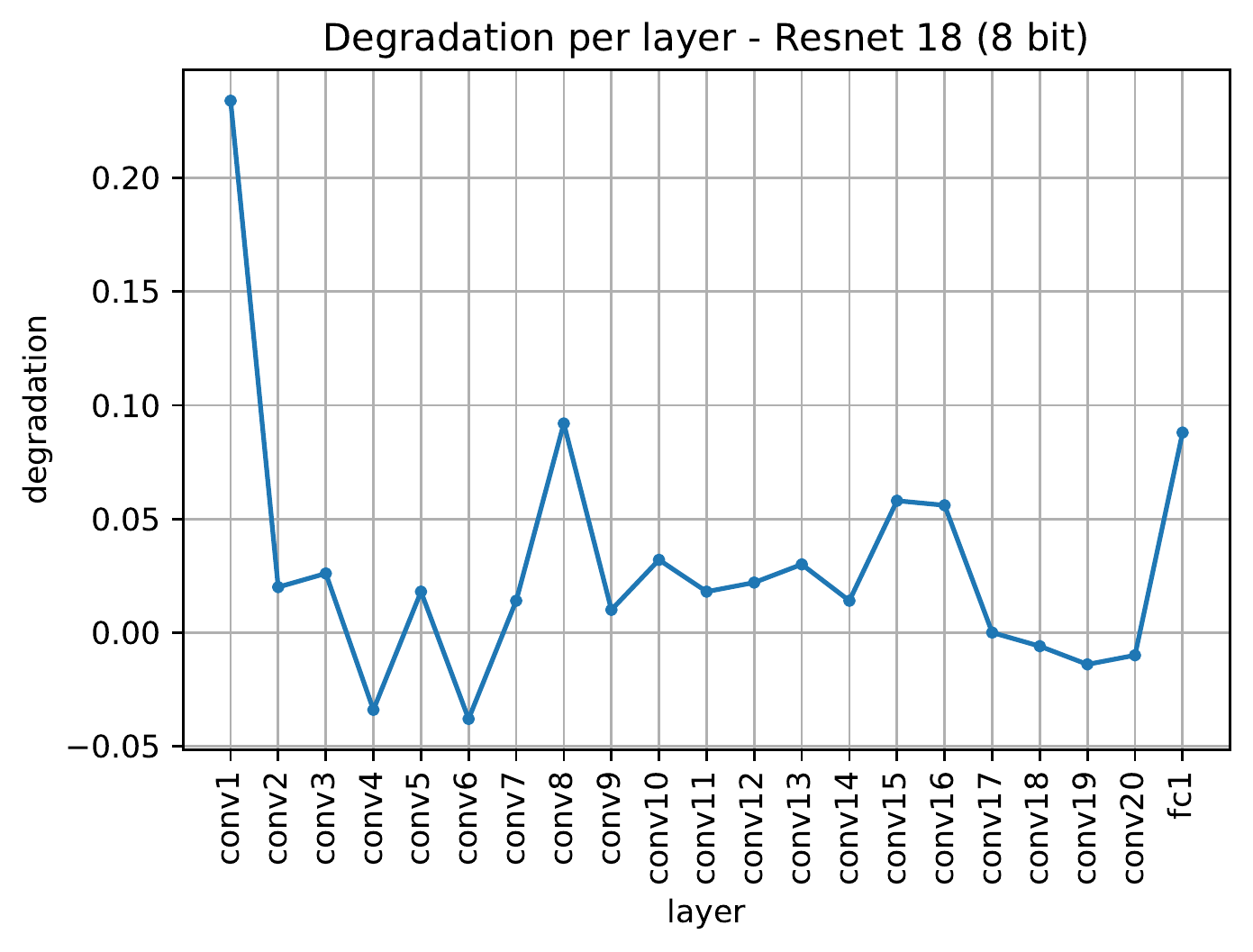}
            \caption{ResNet18 on ImageNet-1K validation set}
            \label{fig:resnet18_analysis}
        \end{subfigure}
    \vspace{0.25cm}
	\end{center}
	\vskip -0.2in
	\caption{\subref{fig:anaylsis_building_block} The building block of our layer-wise analysis is the creation of a network with only a single quantized layer $j$. The input and output of the layer is quantized as well as the weights. The rest of the network is kept at full precision.
		\subref{fig:resnet18_analysis} Layer-wise analysis applied to Resent-18 . This breakdown gives insight into which layers preform poorly under quantization}
	\label{fig:layer_contribution_analysis}
\end{figure}

\section{Introduction}

Quantized neural networks are networks which have their weights and/or activations constrained to low-bit representation, usually 8-bit and below. Neural networks are resource intensive algorithms and quantization is an easy and efficient way to reduce both the computational and memory load of deployed networks.
In recent years, quantization of deep neural networks has become ubiquitous: indeed, just 3 years ago the frontier for quantization was 8-bits for both activations and weights~\cite{jacob2017}, while state-of-the-art works operate at 4-bits and below~\cite{banner2018posttraining,uhlich2019mixed,fang2020nearlossless,jain2019trained,stock2019bit,dong2019hawqv2,wang2018haq, choukroun2019mmse}.

An existing gap in the literature is tools for quantization post-mortem. Once a quantization method has been applied it is evaluated solely on the basis of its degradation with respect to the full-precision network. We claim that an investigation into the sources of quantization degradation is valuable for (a) a better characterization of quantization methods in general and their fail cases in particular, (b) directing future research to address common fail-cases and (c) quick ad-hoc fixes of quantization problems in specific networks.

In this paper we take a step in addressing the above gap. We present \textbf{\textit{layer-wise quantization analysis}} which is based on previously observed properties of quantization: The noise and degradation of individually quantized layers can be summed to get the overall noise and degradation of a fully quantized network~\cite{Zhou2017adaptive}. Our method consists of iterating over the layers of a full precision network and at each step quantizing only a single layer and measuring the resulting noise at the output of the network. Given a quantization method and a network, we can quickly evaluate the layer-wise composition of the overall network degradation. The basic premise of this tool can be summarized by noting that for quantization the whole \textit{is} the sum of its parts. The typical output of such an analysis is shown in \autoref{fig:layer_contribution_analysis}

In \autoref{sec:experiments} we explore the quantization of many neural networks through the prism of layer-wise analysis and observe some interesting phenomena -- most of the degradation is usually caused by few layers. Interestingly, the common practice~\cite{banner2018posttraining,baskin2018nice, esser2019learned,yang2019quantization, dong2019hawq, liu2020posttraining, haroush2019knowledge, nahshan2019loss} of refraining from quantizing the first and last layers doesn't seem warranted as these are seldom the worst offenders. The proposed method makes such analysis easy for a given quantized network and also significantly reduces the level of expertise required in order to achieve successful quantization as it highlights problematic layers on which the effort should to be focused.

In summary, the main contributions of this paper are:

\begin{itemize}
	\setlength\itemsep{1em}

	\item \textbf{Quantization post-mortem} – We develop a simple and efficient way of analyzing quantization degradation by breaking it down to layer-wise contribution. Such analysis can be used for better characterization of quantization methods and gaining insights about the mechanisms by which degradation occurs. Another important aspect of our method is that it enables practitioners without quantization expertise to quickly debug and fix poor quantization by focusing their efforts on specific layers.
	\item \textbf{Improving efficacy of existing methods} – We show that applying some of the existing techniques in a global manner results in worse performance compared to applying them locally. Our method gives an efficient way by which to identify layers that require special treatment and allows targeted application. In \autoref{subsec:resnext-26} we show that we can out-perform existing SoTA techniques by taking this approach.

\end{itemize}

\section{Previous Work}

Quantization has a rich taxonomy but can roughly be divided into two major branches: \textit{Post-training quantization} (PTQ), where the quantization takes place directly on the original full-precision model and \textit{quantization-aware training} (QAT) where training is used to offset the adverse effects of quantization. Post-training methods often try to apply some heuristic to each layer in order to reduce the quantization noise at the output of the layer. Common methods include clipping~\cite{Zhao2019OCS,banner2018posttraining}, channel equalization via inversely-proportional factorization~\cite{meller2019different,nagel2019datafree} and bias-compensation~\cite{bias2019, banner2018posttraining, nagel2019datafree}. A shortcoming of these methods is that they use a local optimization of layer noise in order to achieve global performance optimization. A notable exception is the work of Zhou et. al.~\yrcite{Zhou2017adaptive} which derived a measure of the layer sensitivity to quantization via noise measurement at the output of the network. Their work was the first to decompose overall quantization into layer-wise contributions. In contrast to our work, they use the decomposition in order to establish an analytic correspondence between output noise and degradation in classifiers.
Heuristics for ranking layers based on sensitivity have also been used in the context of network compression where one looks at the optimal bit-assignment per layer. The layer bit-width is determined either explicitly by a measured metric~\cite{Zhou2017adaptive,dong2019hawq,cai2020zeroq} or implicitly via a constrained optimization process~\cite{wang2018haq, elthakeb2018releq} which is resource intensive as it involves re-training the network from scratch. Recently~\citet{uhlich2019mixed}, proposed a simpler method which does not require full re-training. These methods all rank layers based on their contribution of degradation, however, the ranking is also affected by how much can be gained from the compression of the layer. Compared to this methods, our ranking of layers is both more computationally cheap and more direct, not relying on any proxies. Recent works have also assumed individual layer contributions.
\citet{cai2020zeroq} employ a similar method to ours but using KL-divergence instead of noise and network performance. They use their method for mixed-precision quantization in a data-free settings whereas we use our method for analysis of quantization in any settings. \citet{nagel2020adaptive} use a layer-local loss in their method and \citet{hubara2020improving} explicitly optimize over the sum of per-layer loss. Both of these methods are used for post-training optimization and not for analysis. A similar technique to ours was recently used \citet{wu2020integer} to rank layers for exclusion when using partial quantization. This method therefore sidesteps the problem whereas our aim is to both improve full quantization performance and to explore degradation mechanisms. To the best of our knowledge, ours is the first work to use layer-wise quantization as an empirical analysis tool.

\section{Methods}\label{sec:Method}

Here we present our method for layer-wise quantization analysis. We first show how to get the individual degradation of each layer. Then we show that, under practical settings, these contributions are mostly independent and thus can be summed to arrive at the overall network degradation when all layers are quantized. We discuss some limitations of our approach in \autoref{subsec:QFT} and \autoref{sec:bias_shift}.

\subsection{Layer-Wise Quantization Analysis}\label{subsec:layer-wise}

Consider a neural network $L$ with $n$ layers. We topologically sort the layers and enumerate them $l_1,\dots,l_n$. A layer is considered to consist of a typical operation (e.g. convolution, fully-connected, pooling) followed by a non-linearity (e.g. ReLU).
The basic step of our method is the creation of a network with only a single layer, $l_j$, being quantized while the rest of the network is kept at full precision. The quantization of a single layers entails the following: the layer's weights are quantized as well as it's input and output activations. The output of the layer is then de-quantized and propagated onward. See \autoref{fig:anaylsis_building_block} for an illustration.

We use $y$ for the full-precision network output and $\hat{y_j}$ for the output of the network with a quantized layer $l_j$.
The full method consists of iterating over all layers and computing $\hat{y_j}$ for each layer. The evaluation is done over a dataset $D$. In this work we consider two metrics: the $\ell_2$ norm of the quantization noise $||y-\hat{y_j}||^2_2$ and the performance degradation $H_D(y) - H_D(\hat{y_j})$. Where $H_D(y)$ is the metric of the task performed by the network (e.g. top-1 accuracy for a classification network).
The full method is described more formally in \autoref{alg:layer-wise-analysis}.
\begin{algorithm}[!htb]
	\caption{Description of our method for measuring the individual contribution of each layer. For each layer we measure both the noise and degradation resulting from quantization of that layer.}
	\label{alg:layer-wise-analysis}
	\begin{algorithmic}
		\STATE {\bfseries Input:} Dataset $D$, Evaluation function $H$, network $L$ composed of $n$ layers $l_1, l_2, \dots, l_n$
		\STATE {\bfseries Output:} \{$\mathit{noise_i} | i \in \{1 \dots n \}$\}, \{$\mathit{perf_i} | i \in \{1 \dots n \}$\}
		\FOR{$i=1$ {\bfseries to} $n$}
		\STATE	$\mathit{noise_i} = \frac{1}{|D|}\sum\limits_{x \in D}{||y-\hat{y_i}||^2_2}$
		\STATE	$\mathit{perf_i} = H_D(y) - H_D(\hat{y_i})$
		\ENDFOR
	\end{algorithmic}
\end{algorithm}

\begin{figure}[!htb]
	\vskip 0.2in
	\begin{center}
		\includegraphics[width=1.0\linewidth]{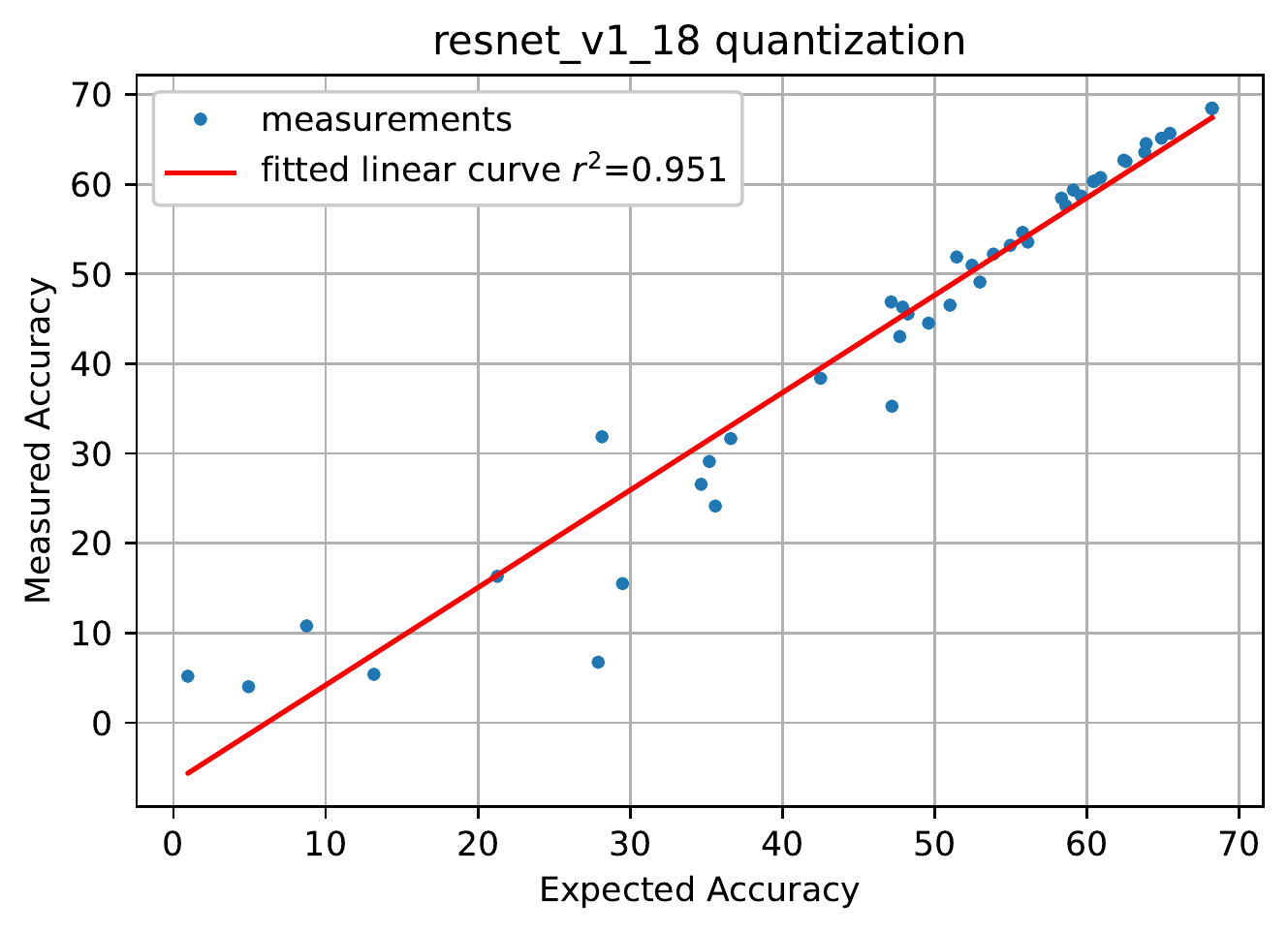}
	\end{center}
	\vskip -0.2in
	\caption{Expected accuracy vs the measured accuracy. Expected degradation is the sum of individual layer degradations. Measurements were done on ImageNet-1K validation set. Each data point is a different quantization of ResNet18. To induce large degradation we quantize a random subset of layers weights to 4 bits while the rest of the layers are quantized to 8 bits. Activations are always quantized to 8 bits. Moving more layers to 4 bits results in larger degradation. We see that the additivity assumption holds well even for very high degradation.}
	\label{fig:degredation addtivity}
\end{figure}

Measuring the individual degradation of each layer is valuable in itself as it immediately shows which layers produce high degradation. It is not clear that we can say something meaningful about the overall network degradation without taking into account the interaction of quantized layers with each other, however,~\citet{Zhou2017adaptive} previously established analytically that the overall noise in the output of a network can be approximated by summing together the output noises of individually quantized layers. The basic assumption underpinning their analysis is that (a) the quantization noises of each layer are independent, (b) quantization noises are small compared to activations and (c) quantization noise has zero mean. In practical settings, these assumptions hold well. Empirically,~\citet{Zhou2017adaptive} show their results hold for noise when only the weights of the network are quantized, however, it is easy to see that their analytical analysis will hold also for the case when activations are also quantized and that the additive nature of noise should extend to degradation as well. To confirm this, we compare the degradation from the sum of individual layers contribution with the measured degradation when quantizing the whole network. The results are shown in \autoref{fig:degredation addtivity}. We see that additivity holds well into the $\sim$10\% degradation range which is much larger than the permissible degradation in practice.
In BRECQ~\cite{anonymous2021brecq} it is suggested that the inter-layer dependence is not negligible. We note that they showed this when quantizing to 2-bits. Other recent works~\cite{Wang_2020_ICML, nagel2020adaptive, hubara2020improving} show good results even when assuming inter-layer independence.

\section{Experiments}\label{sec:experiments}

In this section we experiment with layer-wise quantization and show its two main benefits -- analysis of degradation errors and improving network quantization by applying layer-specific fixes.
We first analyze the results of applying layer-wise analysis to a variety of ImageNet~\cite{ILSVRC15} classifiers. We then show how the quantization of a given network can be improved using the layer-wise breakdown as a compass to pinpoint us to problematic layers.

Unless otherwise stated, we follow the post-training quantization scheme presented in~\cite{jacob2017}. Per-layer quantization is used for both weights and activations. Batch Normalization~\cite{ioffe2015batch} layers are folded into the previous layers. While these are the settings of our experiments, our method is general and can be used both in channel-wise and in group-wise settings.

\subsection{Analyzing ImageNet Classifiers}\label{subsec:analysis}

\begin{figure*}[!htb]
	\vskip 0.2in
	\begin{center}
		\includegraphics[width=0.44\linewidth]{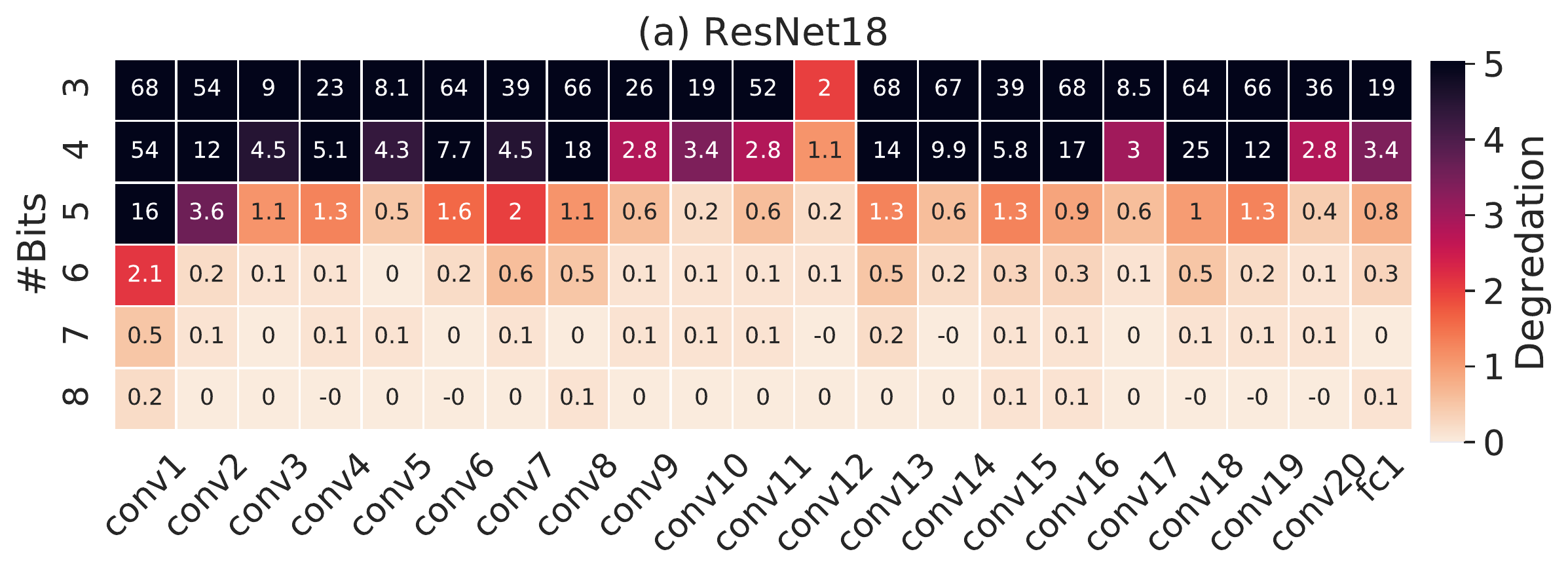}
		\includegraphics[width=0.55\linewidth]{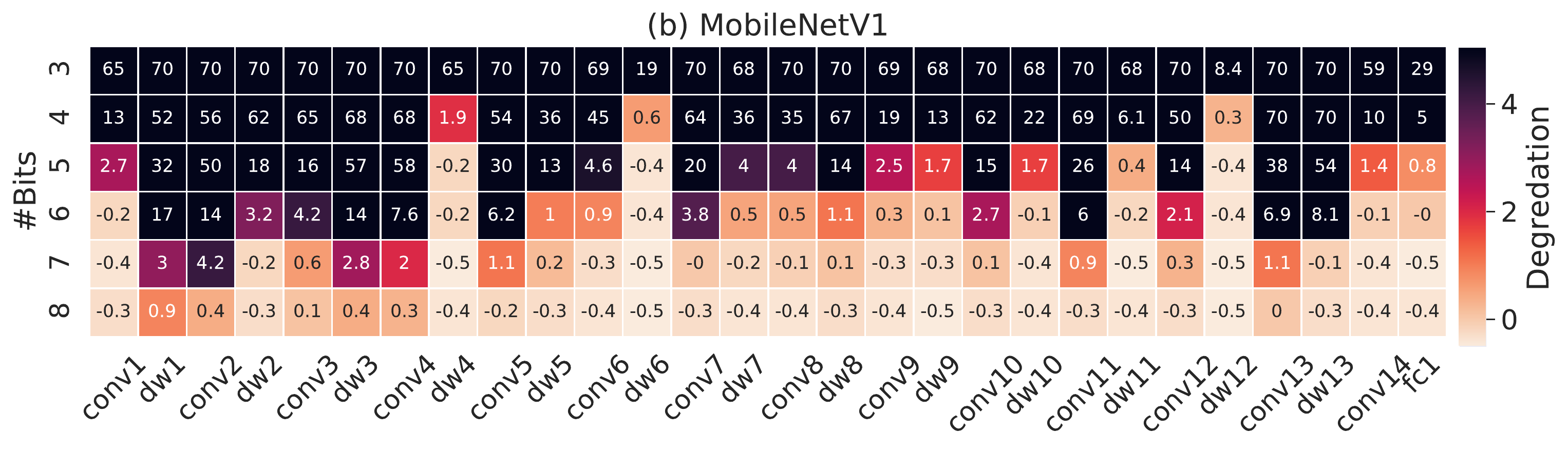}
		\includegraphics[width=1.0\linewidth]{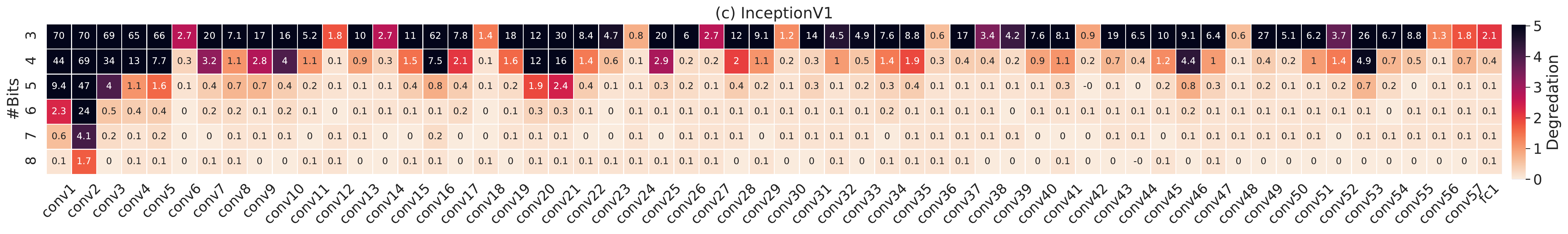}
		\includegraphics[width=1.0\linewidth]{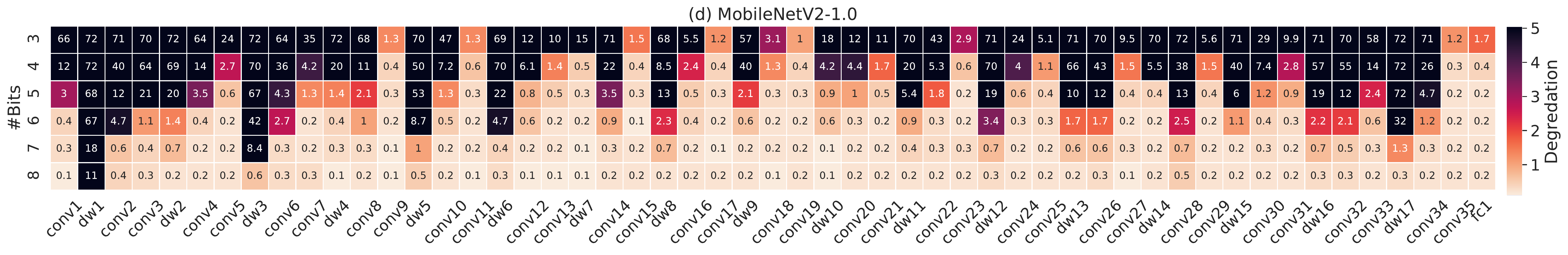}
		\includegraphics[width=1.0\linewidth]{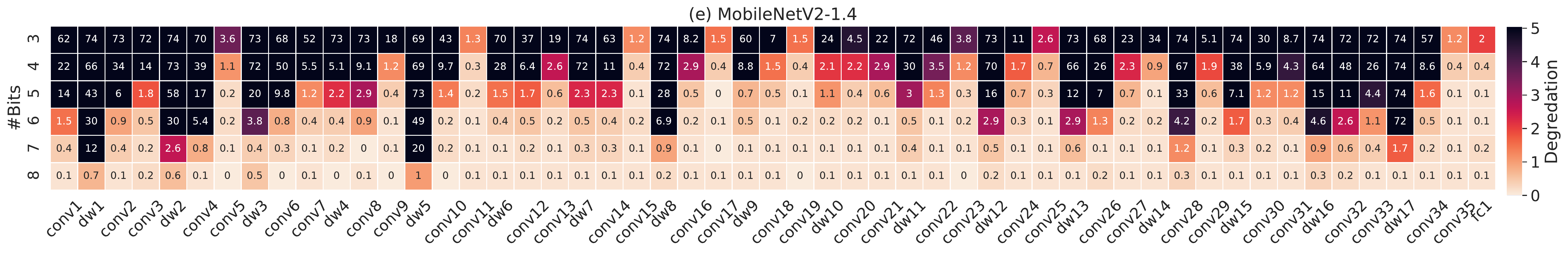}
	\end{center}
	\vskip -0.2in
	\caption{Layer-wise analysis of common networks. Each row is a single run of the tool with a different bit-width selected for the weights. Each column represents a layer of the network. The layers are topologically sorted. The annotation in each cell is the absolute top-1 degradation of the network when only that layer is quantized to a specific bit width. The color bar is saturated at degradation of 5\% for better visibility.}
	\label{fig:Section3_Fig1}
\end{figure*}

We apply our layer-wise analysis to several popular ImageNet classifiers: ResNet18~\cite{he2015deep}, MobileNet-V1-1.0~\cite{howard2017mobilenets}, Inception-V1~\cite{szegedy2014going}, ResNext26~\cite{Xie2016}, MobileNet-V2-1.0 and MobileNet-V2-1.4~\cite{s2018mobilenetv2}. For each network, we analyze the degradation when quantizing each layer's weights between 3-8 bits. Activations are quantized to 8-bits. Degradation is measured over the full ImageNet validation set. The results are shown in \autoref{fig:Section3_Fig1}.

\begin{figure}[!htb]
	\vskip 0.2in
	\begin{center}
		\includegraphics[width=1.0\linewidth]{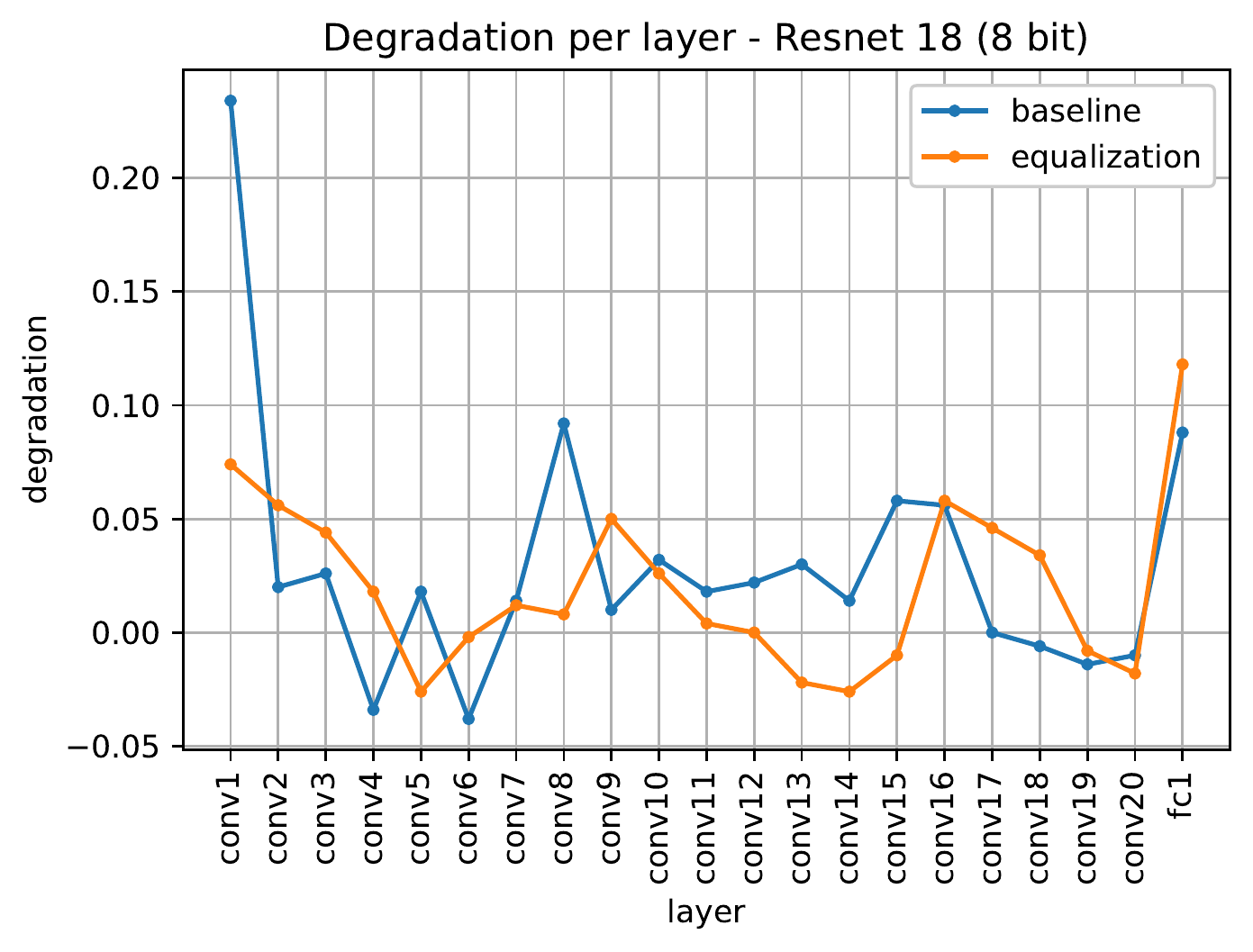}
	\end{center}
	\vskip -0.2in
	\caption{ResNet18 (ImageNet-1K) layer-wise degradation under equalization. The network architecture and full precision output is identical to the baseline model. The results however differ.}
	\label{fig:equalization}
\end{figure}

Looking at the results, we can make several observations. We see that some layers are very sensitive to quantization while others can be quantized to as little as 3-bits with little degradation. A natural question is whether a layer's sensitivity is determined by the network's architecture or is it the result of a specific training. When comparing MobileNet-V2-1.0 and MobileNet-V2-1.4 we see that they have almost identical profiles. These networks share the same architecture but differ in hyper-parameters and are the results of two different training sessions. This would seem to indicate that, under the same training scheme, the sensitivity of layers to quantization is strongly tied to network architecture. To examine whether the results will be similar even under different training settings, we apply Two-Step Equalization~\cite{meller2019different}. Under equalization the network's output and architecture remain unchanged while the network parameters are changed. This simulates a different training scheme while controlling for the network's performance and architecture. As we can see in \autoref{fig:equalization}, the equalized model has a different profile than the non-equalized baseline. This suggests that the training process also plays a role in the network's noise profile under quantization. The question of how much the quantization profile is affected by architecture vs training has important implications. Models in deployment are continuously trained and quantized. Knowing that repeated training session don't require changing quantization schemes can greatly simplify quantization flows, especially those that involve lengthy fine-tuning session -- one should always focus on the layers that do the most harm.
It is interesting to note that many works~\cite{banner2018posttraining,baskin2018nice, esser2019learned,yang2019quantization, dong2019hawq, liu2020posttraining, haroush2019knowledge, nahshan2019loss} refrain from quantizing the first and last layers since it seems to be a commonly held belief that they are harder to quantize. We find that for classifiers the last layer is generally indifferent to quantization and with the exception of ResNet, the first layer is also not the most sensitive layer. Another salient feature is that the depthwise layers in the MobileNet architecture are more sensitive to quantization compared to standard conv2d layers, with the first depthwise layer being particularly finicky.

\subsection{Improving Quantization}\label{subsec:resnext-26}

\begin{figure}[!htb]
	\vskip 0.2in
	\begin{center}
		\includegraphics[width=1.0\linewidth]{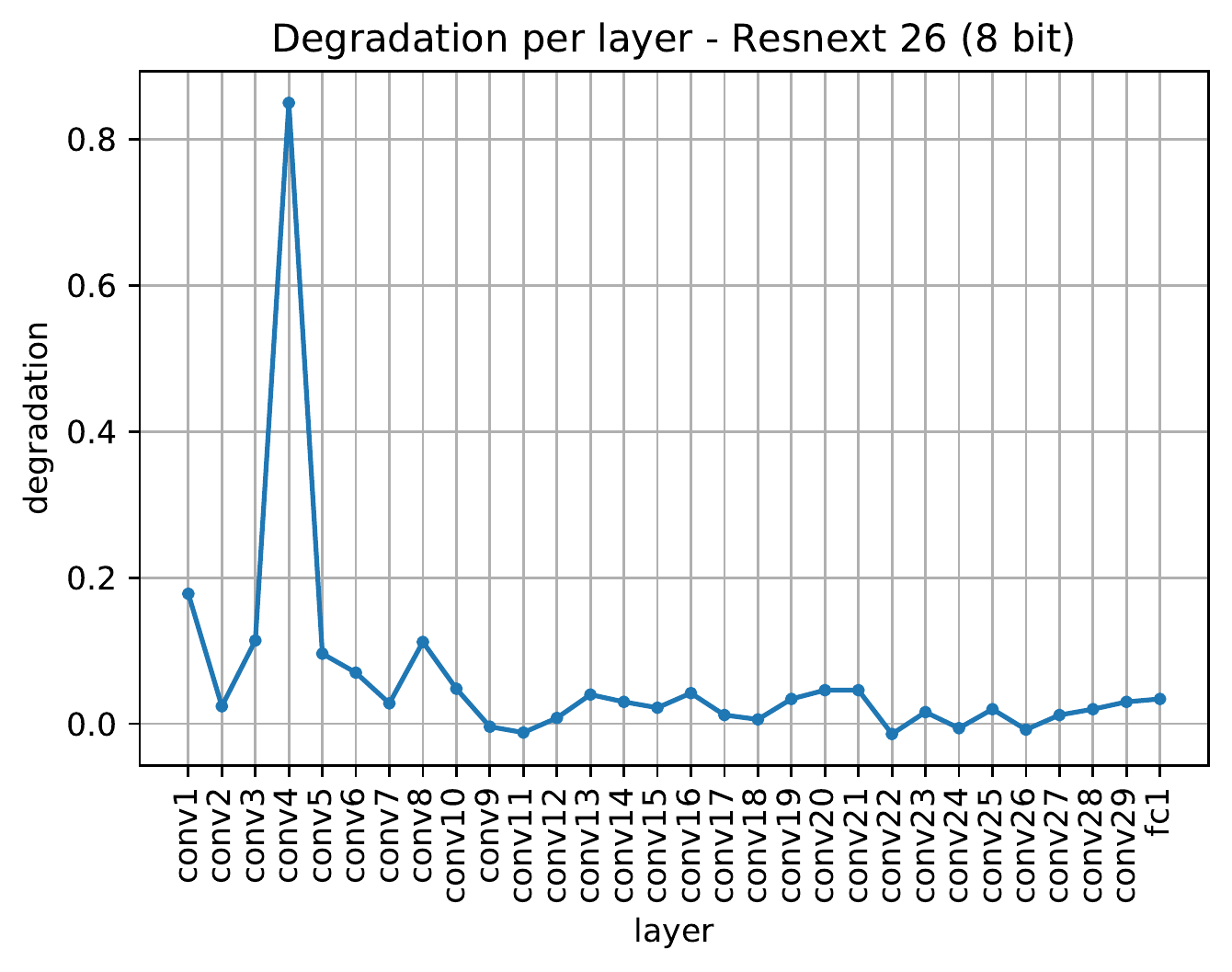}
	\end{center}
	\vskip -0.2in
	\caption{Layer-wise analysis of ResNext26 on ImageNet-1K validation set. Most of the degradation is caused by conv4}
	\label{fig:resnext26_analysis}
\end{figure}

\begin{figure}[!htb]
	\vskip 0.2in
	\begin{center}
		\includegraphics[width=1.0\linewidth]{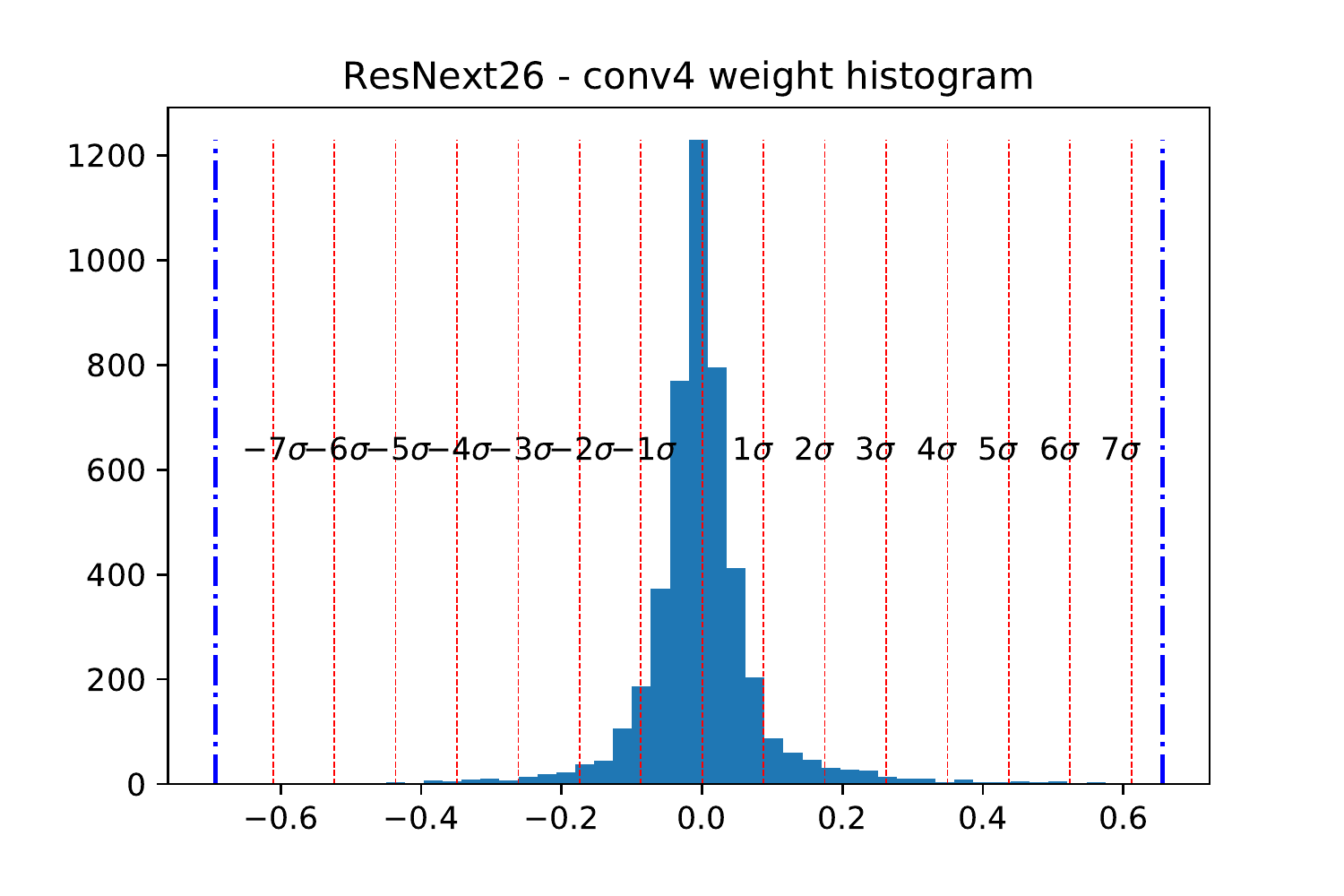}
	\end{center}
	\vskip -0.2in
	\caption{The weight distribution of conv4 in ResNext26. The quantization range is decided by the minimum and maximum values denoted by the dash-dot lines in the figure. We see there are outliers which drastically increase the dynamic range.}
	\label{fig:resnext26_conv4_histogram}
\end{figure}

In this section we show how layer-wise analysis can be leveraged to improve the quantization of a network. To that end, We choose ResNext26~\cite{Xie2016} which has a $>$1\% degradation even when using SoTA post-training techniques such as Equalization~\cite{meller2019different} and IBC~\cite{bias2019}. The results of running the analysis on the network are shown in~\autoref{fig:resnext26_analysis}. We see that almost all of the degradation stems from a single layer -- conv4. Since 8-bit quantization usually gives good results, it is expected that layer exhibiting large degradation will be rare. We now examine the problematic layer. Since quantization noise is the result of the trade-off between representation range and precision, excess degradation is commonly due to large ranges of kernel values. In~\autoref{fig:resnext26_conv4_histogram} we plot the weights histogram of conv4 and indeed we see significant outliers. At this juncture, it is reasonable to try and use a standard weight clipping technique such as SAWB\footnote{The results we report are not sensitive to the type of clipping used provided it is aggressive enough to handle the outliers}~\cite{choi2018bridging} on conv4 which results in near lossless quantization. We emphasize that this result is possible due to the \textit{local} application of clipping, as applying SAWB globally degrades the network performance even below baseline as shown in \autoref{tab:resnext-quantization}. This example illustrates a general principle: quantization techniques can be beneficial locally but degrade performance when used globally. Since PTQ methods are inherently heuristic, the standard practice of evaluating them only on the merit of global application is bad practice; moreover, compound quantization methods which apply a larger-range of heuristics locally can give much better results.

\begin{table}[t]
	\vskip 0.15in
	\begin{center}
		\begin{threeparttable}
			\caption{Top-1 Accuracy for ResNext26 on ImageNet-1K validation set under different post-training methods for 8-bit quantization. The standard approaches still show high degradation.
				By using our layer-wise analysis we identify the problem and manage to get better results.}
			\begin{tabular}{@{}c c@{}}
				\toprule
				\textbf{Method}                   & \textbf{Top-1 Accuracy} \\
				\midrule
				Full Precision                    & 76.08                   \\
				\midrule
				Naive Quantization                & 74.28                   \\
				Equalization + IBC                & 75.00                   \\
				\midrule
				SAWB globally                     & 71.88                   \\
				\textbf{SAWB conv4\tnote{$\psi$}} & \textbf{75.91}          \\
				\bottomrule
			\end{tabular}
			\begin{tablenotes}\footnotesize
				\item[$\psi$] Methods using our layer-wise breakdown
			\end{tablenotes}

			\label{tab:resnext-quantization}
		\end{threeparttable}
	\end{center}
	\vskip -0.1in
\end{table}

\subsection{Quantization-Aware Training}\label{subsec:QFT}

\begin{figure}[!htb]
	\vskip 0.2in
	\captionsetup[subfigure]{aboveskip=-1pt,belowskip=-1pt}
	\begin{center}
	    \begin{subfigure}{0.5\textwidth}
            \includegraphics[width=1.0\linewidth]{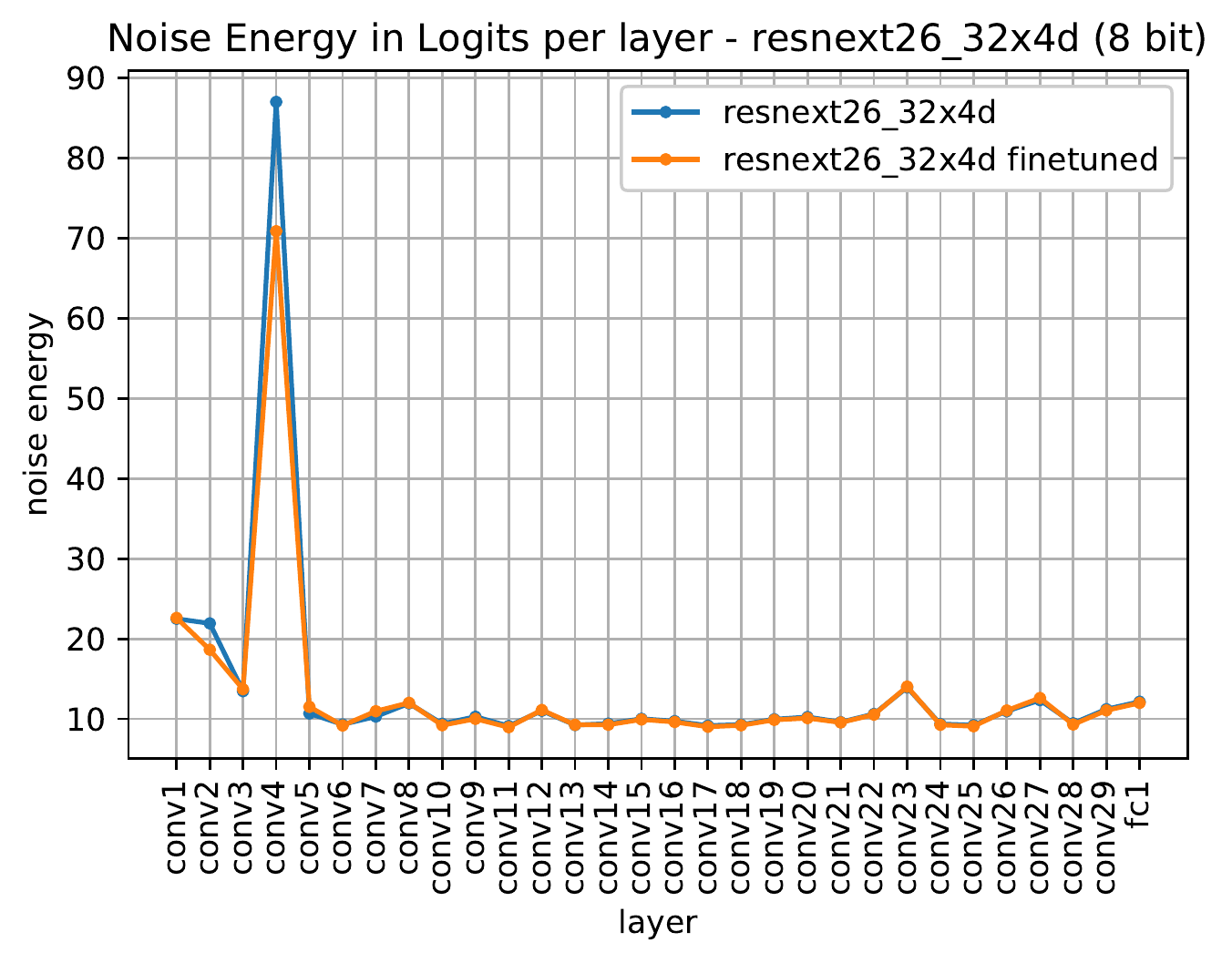}
            \caption{}
            \label{fig:resnext26_finetune_noise}
        \end{subfigure}
        \begin{subfigure}{0.5\textwidth}
            \includegraphics[width=1.0\linewidth]{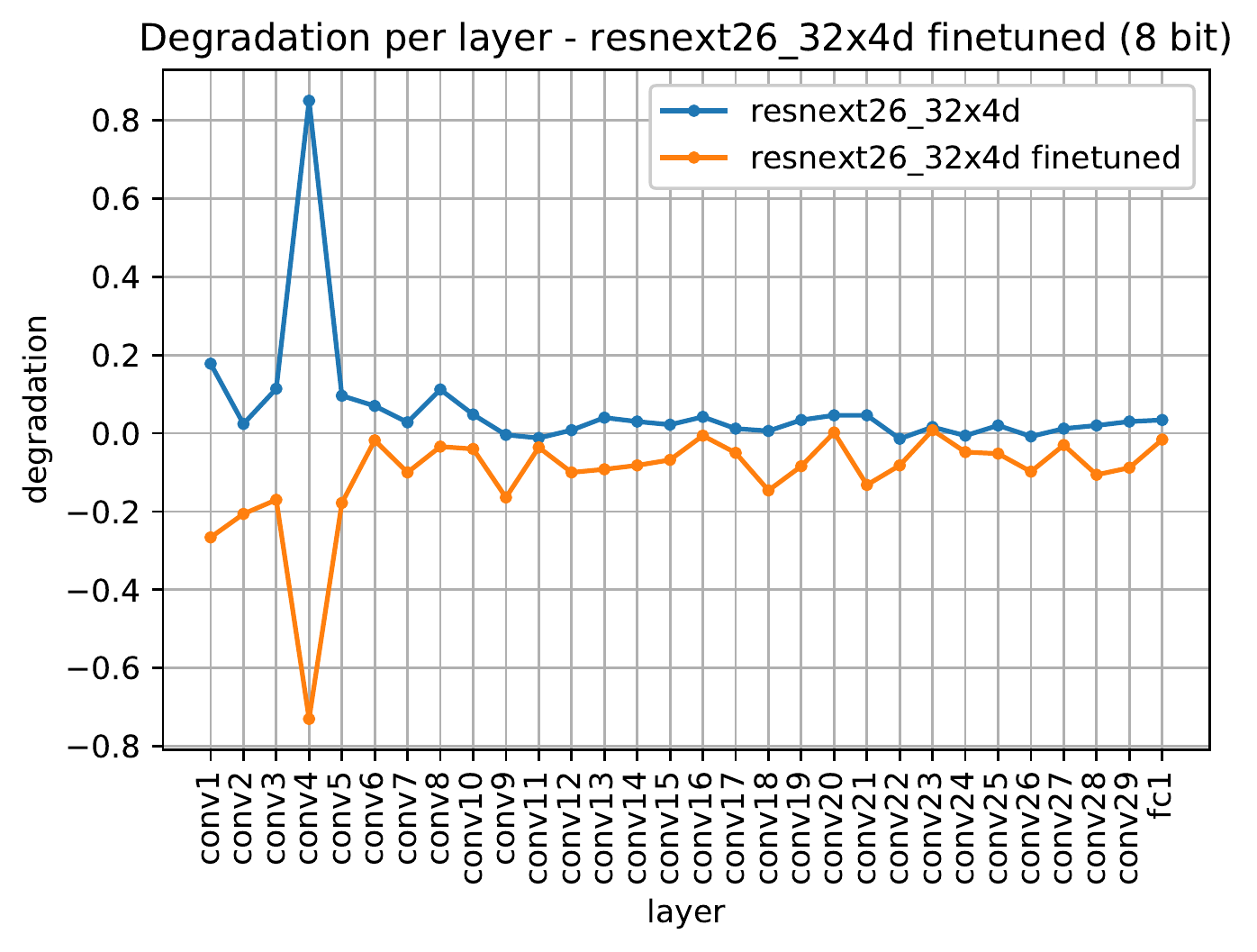}
            \caption{}
            \label{fig:resnext26_finetune_deg}
        \end{subfigure}
	\end{center}
	\vskip -0.1in
	\caption{Layer-wise analysis of ResNext-26 on ImageNet-1K validation set before and after fine-tuning. The usual correlation between noise and degradation breaks down. \subref{fig:resnext26_finetune_noise} Noise energy has decreased in conv4 after fine-tuning, but is still high. The rest of the noise profile is similar. \subref{fig:resnext26_finetune_deg} The Fine-tuned network is showing improved accuracy when a layer is quantized. Using the full-precision network after fine-tuning as the reference network is no longer appropriate.}
	\label{fig:resnext26_finetune}
\end{figure}

Our method was designed to be used in a post-training quantization setting. Networks which undergo quantization-aware fine-tuning~\cite{jacob2017} via straight-through estimation can't be used with our method due to the fact that the full-precision network is no longer the appropriate reference. It is, however, still of interest to examine the results. We continue with the same network as in the previous section. The optimization target of the fine-tune process is \textit{the quantized network} and so it doesn't act to reduce the noise in the full-precision network but rather to compensate for it in the quantized network. This is illustrated in \autoref{fig:resnext26_finetune} where we see that the fine-tuned network has almost the same noise profile as the original network, however, since the network was trained in the presence of noise, the performance of the fine-tuned network is \textit{improved} when noise is introduced to the network by quantization.

\begin{figure}[!htb]
	\captionsetup[subfigure]{aboveskip=-1pt,belowskip=-1pt}
	\begin{center}
	    \begin{subfigure}{0.5\textwidth}
            \includegraphics[width=1.0\linewidth]{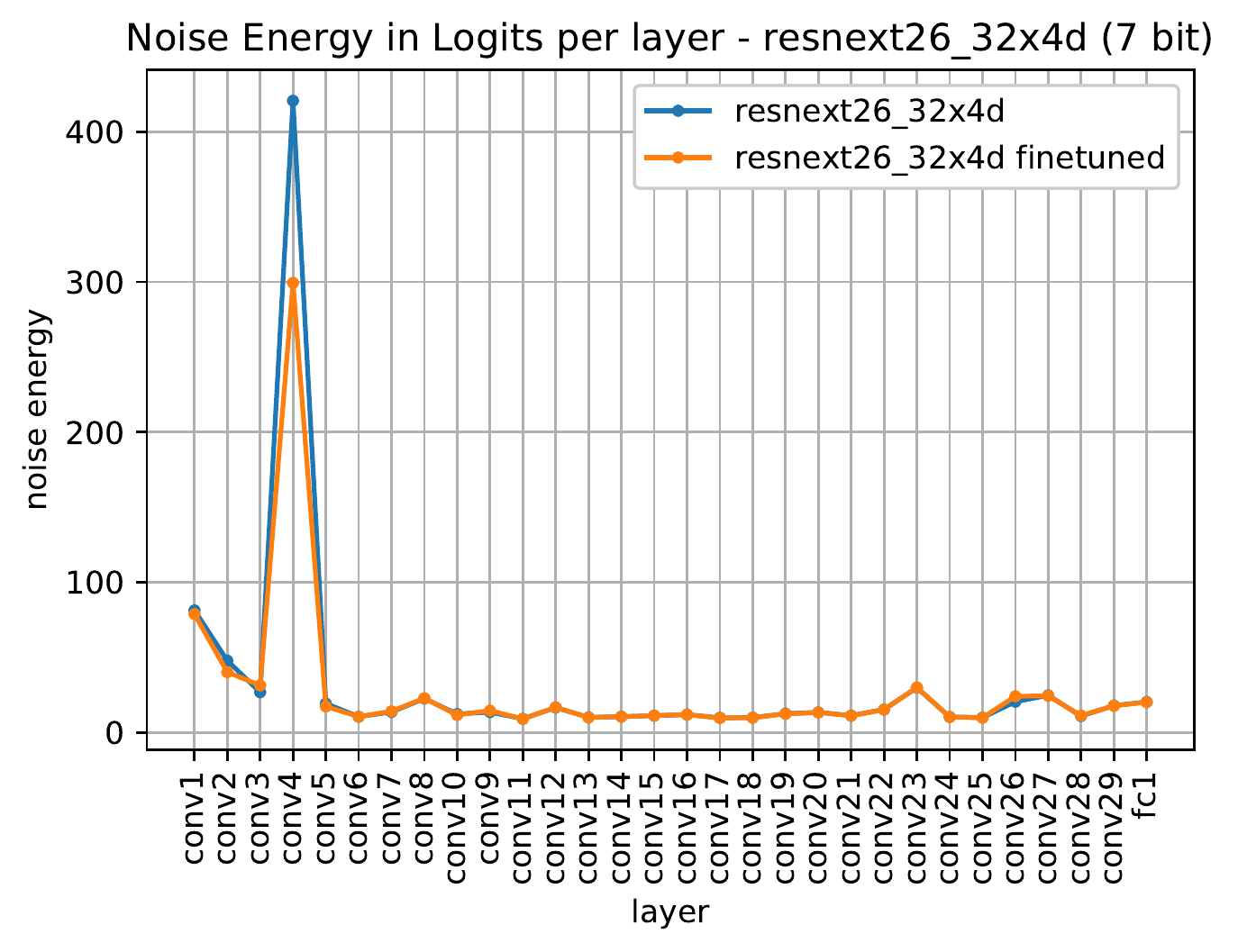}
            \caption{Layer-wise analysis of noise}
            \label{fig:resnext26_finetune_7bit_noise}
        \end{subfigure}
        \begin{subfigure}{0.5\textwidth}
            \includegraphics[width=1.0\linewidth]{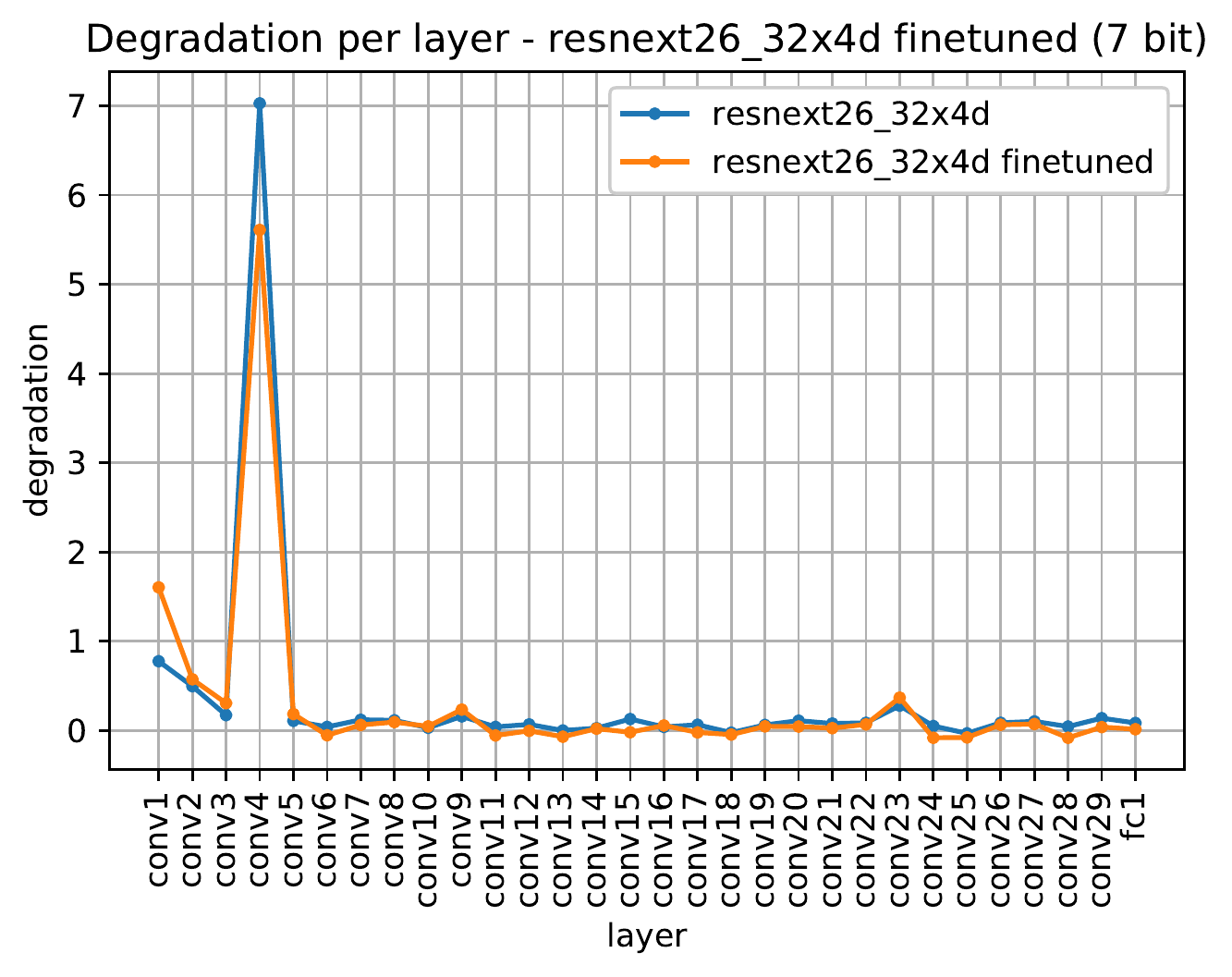}
            \caption{Layer-wise analysis of degradation}
            \label{fig:resnext26_finetune_7bit_deg}
        \end{subfigure}
	\end{center}
	\vskip -0.1in
	\caption{layer-wise analysis of ResNext-26 before and after fine-tuning to 8-bit quantization. The network is quantized 7 bits when performing the analysis to make the quantization noise independent of the 8-bit noise used for fine-tuning. We see that the correlation between noise and degradation is restored.}
	\label{fig:resnext26_finetune_7bit}
\end{figure}

The problem stems from the fact that the full-precision network and the optimization target of the training process are no longer one and the same.
Our type of analysis only works well when the reference against which we compare the quantized network is the optimization target.
Another way of expressing the above is that when the network is trained in the presence of a specific noise profile then the assumption of noise independence no longer holds. Thus, it is still possible to use our method to analyze fine-tuned models, but we must use quantization noise that is independent of the network training. This is shown in \autoref{fig:resnext26_finetune_7bit} when we take the full-precision model fine-tuned to 8-bit quantization and use 7-bits quantization instead. This above doesn't mean that layer-wise analysis is of no use when using quantization-aware training. The fact that the fine-tuning only compensates but does not reduce the noise, means that performing an analysis prior to lengthy training is still beneficial - problematic layers can be recognized and dealt with prior to training for improved performance.

\subsection{Bias Shift}\label{sec:bias_shift}

One of our underlying assumptions for noise additivity is that quantization noise has zero mean. Previous works~\cite{bias2019,nagel2019datafree} have shown that layers with small kernels (e.g. depthwise layers) receive a mean activation shift~\cite{bias2019} due to quantization. For these networks there is strong interaction between noises from adjacent layers, skewing the result of layer-wise analysis. For example, using layer-wise analysis on MobileNets~\cite{howard2017mobilenets,Sandler2018} (\autoref{fig:Section3_Fig1}) which have many depthwise layers, shows that the linear additivity of quantization noise no longer holds. We note that we can still apply layer-wise analysis on MobileNets (\autoref{subsec:analysis}) for finding the individual degradation of layers, however, the sum of these individual contributions will under-estimate the overall network degradation.

\begin{figure}[!htb]
	\vskip 0.2in
	\begin{center}
		\includegraphics[width=1.0\linewidth]{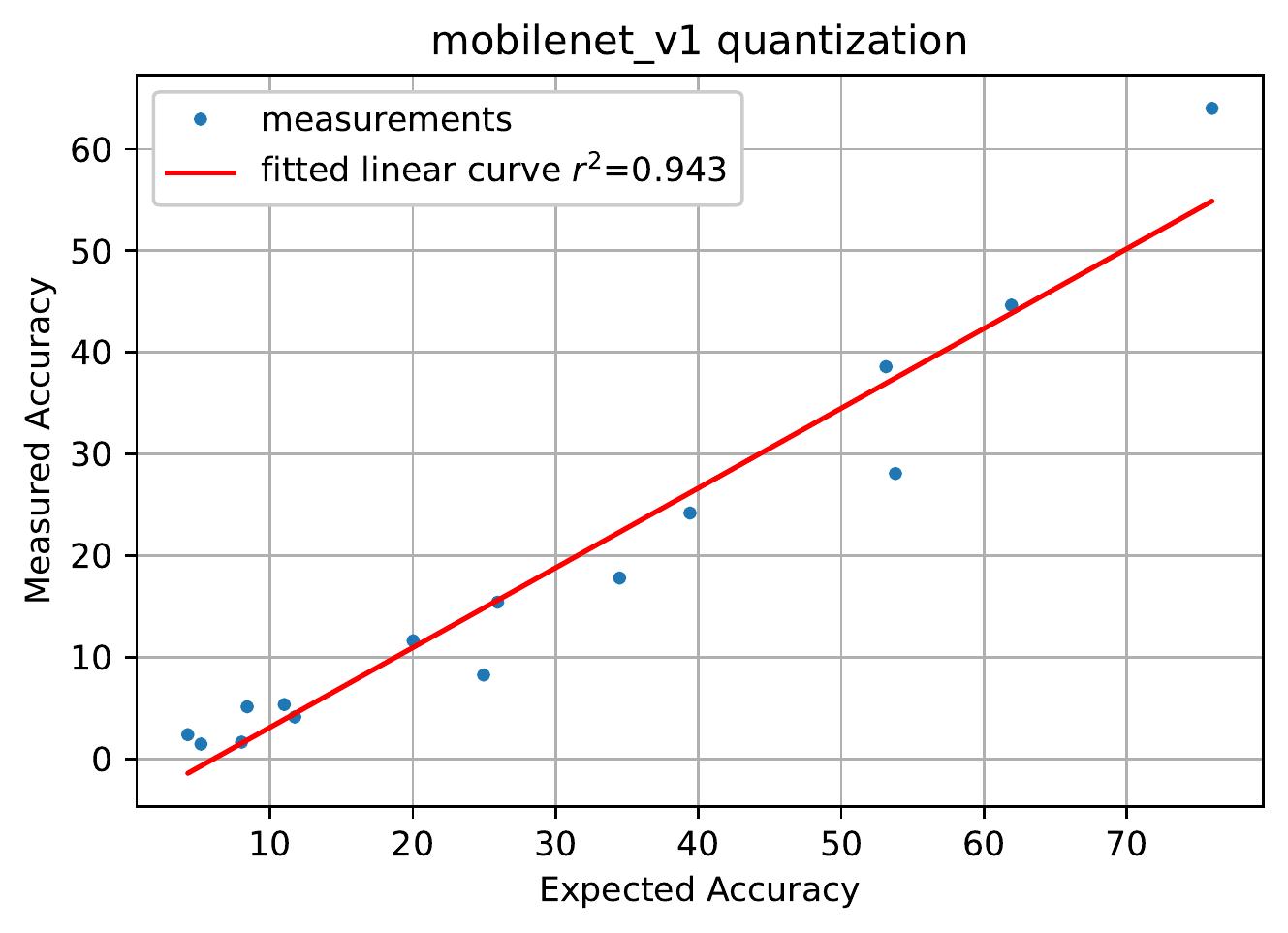}
		\includegraphics[width=1.0\linewidth]{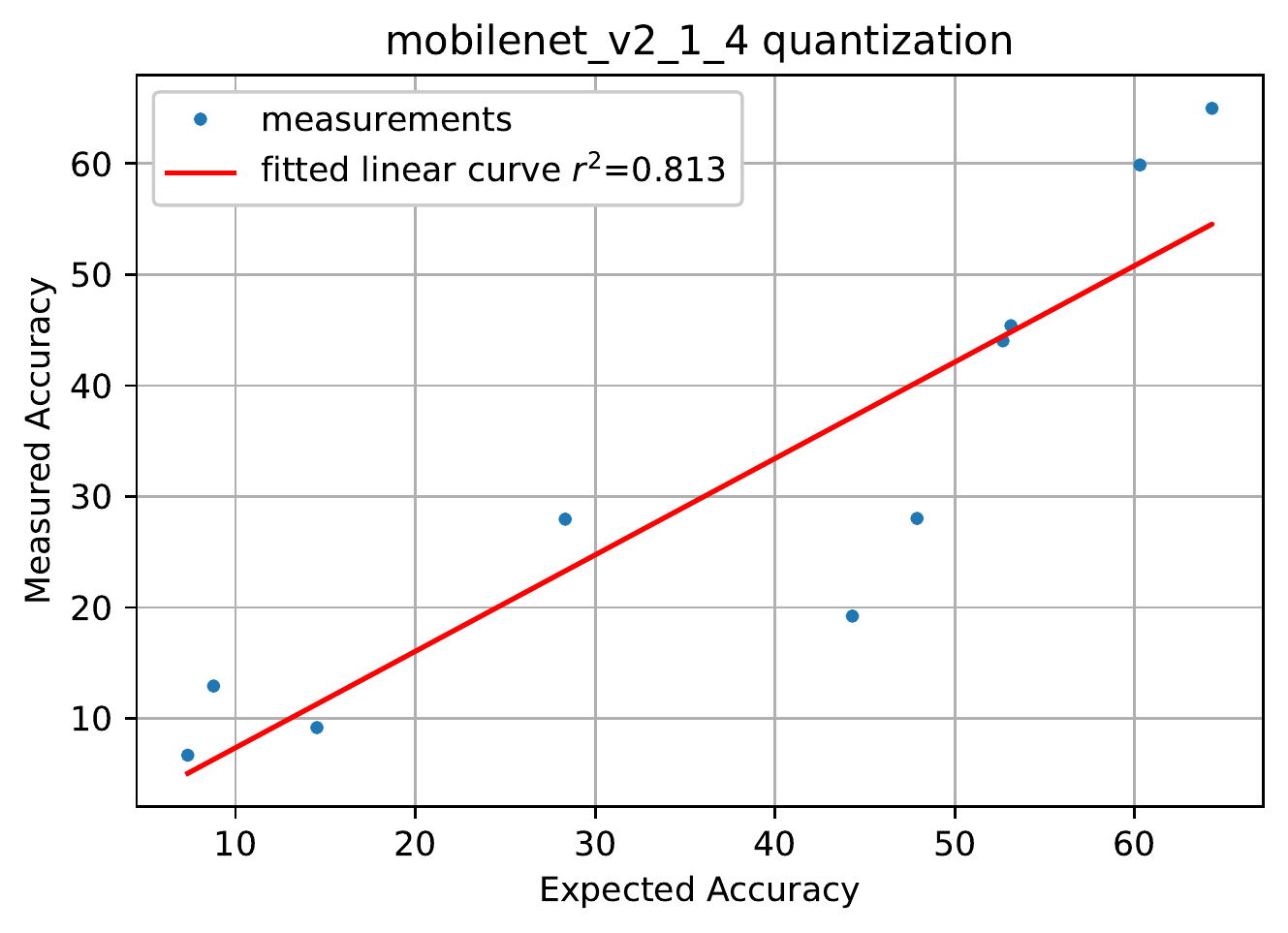}
	\end{center}
	\vskip -0.2in
	\caption{MobileNets expected degradation vs. measured. Correlation is lower due to bias shift. Measurements were done on ImageNet-1K validation set and were taken similarly to \autoref{fig:degredation addtivity}}
	\label{fig:degredation addtivity_mobilenet}
\end{figure}

\section{Discussion and Future Directions}\label{sec:discussion}

In this work, we presented a method for layer-wise analysis of post-training quantization methods. The use-case we presented in \autoref{subsec:resnext-26} illustrates a general issue with advanced post-training quantization methods  -- they often have some hyperparameter (e.g. clipping value) which is selected based on greedy optimization. Values that are optimal for certain layers can be detrimental to others to the point where overall network performance is degraded. Our tool directly measures the contribution of the layer to the overall degradation, taking into account both the layer location in the network as well as the whole network architecture. Once problematic layers have been identified it is easy to apply existing methods in a more precise fashion. We further claim that the difference between global and local application of quantization methods is symptomatic of a larger issue which is the one-dimensional evaluation of quantization methods. As shown in \autoref{subsec:analysis} using layer-wise analysis gives a more nuanced evaluation of the performance of quantization methods, allowing disambiguation between architecture-specific effects and general qualities of the quantization method. We believe that it is important for future works to give a layer-wise breakdown of their method that will show exactly from which layers the improvement in performance is achieved. This will allow both better understanding and create a richer design space for compound quantization methods. Our method is a first step in the direction of establishing better analysis tools for quantization which we hope will serve as a framework for future works.

\bibliographystyle{icml2020}
\bibliography{LayerQuantizationAnalysis}

\end{document}